\documentclass[letterpaper, 10 pt, conference]{ieeeconf}  

\IEEEoverridecommandlockouts                              

\usepackage{tabularx}
\usepackage{verbatim}
\usepackage{amsfonts}
\usepackage{amsmath}
\usepackage{arydshln}
\usepackage{units}
\usepackage{graphicx}
\usepackage{subcaption}
\usepackage{tikz}
\usepackage{float}
\usepackage{multirow}
\usepackage{makecell}
\usepackage{array}
\usepackage{xcolor}
\usepackage{hyperref}
\usepackage{comment}
\usepackage{caption}

\usepackage[natbib=true,style=numeric, sorting=none,maxcitenames=40,mincitenames=4,giveninits=false, citestyle=numeric-comp]{biblatex}
\usepackage{graphicx}
\graphicspath{ {./Figures/} }
\usepackage[ruled, lined, commentsnumbered, longend]{algorithm2e}

\definecolor{commentcolor}{RGB}{110,154,155}   
\newcommand{\PyComment}[1]{\ttfamily\textcolor{commentcolor}{\# #1}}  
\newcommand{\PyCode}[1]{\ttfamily\textcolor{black}{#1}} 

\def\lowcomma{_{\textstyle,}}
\def\lowdot{_{\textstyle.}}

\title{\LARGE \bf
Multimodal Visual-Tactile Representation Learning through Self-Supervised Contrastive Pre-Training
}
\author{
    \begin{tabular}{c}
        Vedant Dave\textsuperscript{*}, Fotios Lygerakis\textsuperscript{*}, Elmar Rueckert
    \end{tabular} \\
    {\normalsize{Cyber-Physical-Systems Lab at Montanuniversität Leoben, Austria}}
    \thanks{\textsuperscript{*}These authors contributed equally to this work.}
    \thanks{Corresponding Author: \tt\small {vedant.dave@unileoben.ac.at}}
}

\begin{document}

\maketitle
\thispagestyle{empty}
\pagestyle{empty}

\begin{abstract}
The rapidly evolving field of robotics necessitates methods that can facilitate the fusion of multiple modalities. Specifically, when it comes to interacting with tangible objects, effectively combining visual and tactile sensory data is key to understanding and navigating the complex dynamics of the physical world, enabling a more nuanced and adaptable response to changing environments.
Nevertheless, much of the earlier work in merging these two sensory modalities has relied on supervised methods utilizing datasets labeled by humans.
This paper introduces MViTac, a novel methodology that leverages contrastive learning to integrate vision and touch sensations in a self-supervised fashion. By availing both sensory inputs, MViTac leverages intra and inter-modality losses for learning representations, resulting in enhanced material property classification and more adept grasping prediction. 
Through a series of experiments, we showcase the effectiveness of our method and its superiority over existing state-of-the-art self-supervised and supervised techniques.
In evaluating our methodology, we focus on two distinct tasks: material classification and grasping success prediction. Our results indicate that MViTac facilitates the development of improved modality encoders, yielding more robust representations as evidenced by linear probing assessments.
{\footnotesize \url{https://sites.google.com/view/mvitac/home}}

\end{abstract}

\section{INTRODUCTION}
In the realm of robotics, visual perception has traditionally served as a central modality extensively leveraged for acquiring nuanced environmental representations, a role emphasized in a range of studies \cite{10087821, robotics11060139}. However, this approach harbors intrinsic limitations in fully encapsulating the dynamic and intricate state of the surrounding environment \cite{doi:10.1152/jn.00439.2014}. Conversely, tactile sensing excels in delineating fine-grained attributes that are beyond the grasp of visual modalities, effectively capturing the subtleties that evade visual systems. 

Thus, a synergic integration of both visual and tactile modalities offers a more robust and comprehensive world representation, wherein the visual systems predominantly decipher global features while tactile sensing augments this with enriched local feature representations \cite{Calandra2018MoreTA, doi:https://doi.org/10.1002/cphy.c170033}. However, the fusion of these modalities is far from trivial, given the disparate information density each conveys over time. Especially in manipulation task settings, the reliance on vision remains high until the robot engages physically with an object or surface; following this interaction, the tactile modality often becomes the principal source of nuanced data, particularly in scenarios involving occlusion by the robotic arm.

Recent endeavors are increasingly focusing on the efficient fusion of vision and tactile representations to navigate the challenges delineated above \cite{7462208, li2022seehearfeel, 8460494, chen2022visuotactile, kerr2023selfsupervised}. This is further propelled by the significant advancements in self-supervised learning (SSL) approaches, catalyzed by the lack of labeled tactile and visuotactile datasets \cite{kerr2023selfsupervised, yang2022touch, guzey2023dexterity, chen2022visuotactile}. 
In contrast to visual data, where simulations and virtual environments can often provide substantial and rich datasets, tactile data collection requires physical interaction with a wide array of materials and objects to encapsulate a rich diversity of tactile experiences. This not only escalates the complexity but also significantly extends the time needed for data collection. Recognizing these challenges, researchers are turning towards the SSL strategy to leverage unlabeled data, which is easier and quicker to collect \cite{kerr2023selfsupervised, yang2022touch, guzey2023dexterity}. 
SSL facilitates the learning of useful representations from this unlabeled data, potentially accelerating the development of sophisticated visuotactile systems by reducing the dependency on labor-intensive labeled datasets, thus presenting a promising avenue to mitigate the hurdles in tactile data collection. In recent years, contrastive learning (CL) has risen as a prominent subfield of self-supervised learning (SSL), establishing itself as the predominant methodology for pretraining visual encoders \cite{CPC,simCLR, moco, byol, swav, vicreg}.

In this study, we propose a \textbf{M}ulti-modal \textbf{Vi}sual-\textbf{Tac}tile (MViTac) representation learning algorithm for effectively fusing the two modalities. Our methodology facilitates the learning of both intra and inter-modal representations, leveraging the richness of visual and tactile observations. We use two sets of encoders to compute the InfoNCE loss for the two ranges of losses. We employ the within-modality (intra) loss to maximize the agreement of the representations of similar modality instances. Similarly, we employ the across-modality (inter) loss to maximize the similarity of representations of different modalities across the same sample. 
Utilizing the Touch-and-Go \cite{yang2022touch} and Calandra \cite{Calandra2018MoreTA} datasets  for training the encoders, we undertake an exhaustive evaluation across diverse downstream tasks, including material property identifications and robot grasp prediction. We thereby substantiate the efficacy of our approach in transcending the performance benchmarks set by previous state-of-the-art frameworks.

\section{RELATED WORK}
The integration of visual and tactile modalities in robotic systems is a growing research area, with numerous studies delving into its complexities. This section highlights key contributions in this field.

\subsection{Tactile Sensing in Robotics}
Humans can identify physical properties (hardness, roughness, texture) of objects exclusively through tactile interactions~\cite{identifyingobject} and exhibit significant reliance on tactile feedback for grasping and manipulation tasks. In recent years, various methods have been developed for the extraction of tactile information, proving instrumental for robotic applications~\cite{tactilereview2,tactilereview3}. Works such as~\cite{deng2020grasping,graspstability,si2022grasp} have utilised tactile signals to estimate the object grasp stability, while Veiga et al.~\cite{veiga2020grip} extended this idea to independent fingers for detecting local slips. Significant robotics research focuses on manipulation and grasping that integrates object properties and geometry, gripper configurations, and environmental conditions ~\cite{tactovis,li2020review}. Various methodologies like Gaussian Processes~\cite{calandra2015learning,veiga2017tactile}, Movement Primitives~\cite{hogan2020tactile,tacpromps}, Reinforcement Learning~\cite{regrasp,pageflip} are incorporated to obtain the correlation between the manipulation skills and tactile sensations. In recent years,  vision-based optical tactile sensors such as Gelsight \cite{gelsight}, Digit~\cite{lambeta2020digit}, and XELA uSkin~\cite{xelauskin} have gained significant traction due to their ability to provide rich information concerning object geometry, forces, and shear.

\subsection{Vision and Touch in Robotic Manipulation}
Several advancements have been made in the integration of tactile sensing for enhanced grasping and manipulation. Calandra et al. \cite{DBLP:journals/corr/abs-1710-05512} demonstrated that integrating tactile sensing significantly enhances grasping results. Their approach involved training vision-only, tactile-only, and visual-tactile grasp success predictors and selecting the one with the highest success probability. In another study \cite{tactilegym2}, a tactile encoder was trained in simulation across three distinct tasks and managed to achieve zero-shot sim-to-real transfer using a generative adversarial network. Delving deeper into tactile representations, Guzey et al. \cite{guzey2023dexterity} found that utilizing a self-supervised method to learn these representations led to superior results in manipulation tasks. Their approach gathered a dataset of teleoperated, contact-rich, albeit arbitrary, interactions with the environment, which they then used to derive tactile representations. For subsequent tasks, they employed a nearest neighbors approach to recall actions. Meanwhile, the research outlined in \cite{9812019} introduced Visuotactile-RL, a methodology combining visual and tactile feedback for manipulation. They harnessed tactile data sourced from optical sensors and explored two image encoder architectures, namely MultiPath (MP) and SinglePath (SP), complemented by tactics such as tactile gating.

\subsection{Visual-Tactile Joint Representation learning}
In recent studies, the fusion of visual and tactile data has emerged as a focal point in advancing object recognition and manipulation tasks. Li et al. in \cite{li2019connecting} presented a cross-modal prediction system, which addresses the significant scale gap between visual and tactile signals using conditional adversarial networks. The system synthesizes temporal tactile signals from visual inputs and identifies touched object parts from tactile inputs, enhancing the interaction between vision and touch. The authors in \cite{li2019connecting} employ a data rebalancing strategy to prevent mode collapse during GAN training and the inclusion of touch scale and location data in the model. H. Liu et al. \cite{7462208} developed a visual-tactile fusion framework using a joint group kernel sparse coding method to address the weak pairing issue in visual-tactile data samples. Further contribution by H. Li et al. leveraged three distinct attention mechanisms including multi-head self-attentions across different modalities and timelines, proving beneficial in mastering dense packing and pouring tasks \cite{li2022seehearfeel}. S. Luo et al. \cite{8460494} aimed at amplifying cloth texture recognition precision by focusing on shared features across different modalities. They tested their system in robotic tactile exploration tasks for cloth material identification. The studies extended into robot-assisted dressing with a notable focus on garment unfolding, integrating visual-tactile prediction models with reinforcement learning to guide robotic movements effectively during the unfolding process \cite{10185075}. Chen et al. \cite{chen2022visuotactile} introduced the Visuo-Tactile Transformer (VTT) which utilized spatial attention for merging visual and tactile data, demonstrating enhanced efficiency in correlating tactile events with visual cues. 
Kerr et al. \cite{kerr2023selfsupervised} build a self-supervised learning strategy that leverages intra-modal contrastive loss for learning representations in tasks such as garment feature tracking and manipulation. These endeavors collectively underscore the promising trajectory of integrating visual and tactile feedback for refined robotic functionalities in recognition and manipulation tasks.




\section{MULTIMODAL SELF-SUPERVISED LEARNING}
We introduce a cross-modal self-supervised learning approach for learning representations between visual and tactile data, which can subsequently be employed for downstream applications such as material classification and robotic manipulation tasks.
\subsection{Problem Statement}
Given a visual observation $\mathcal{O}_V$ and a tactile observation $\mathcal{O}_T$ both corresponding to the same object, our objective is to find a function $f$ that encapsulates the projections $z_V=f(\mathcal{O}_V$) and $z_T=f(\mathcal{O}_T)$, which lie close to each other in a lower-dimensional latent space $\mathcal{Z}$, $z_V, z_T \in \mathcal{Z}$. The observation spaces $\mathcal{O}_V$ and $\mathcal{O}_T$ are RGB images in $	\mathbb{R}^{H\times W\times 3}$, where $H$ and $W$ are the height and the width of the images.
\begin{figure*}[htb]
  \begin{minipage}[t]{0.45\textwidth}
    \vspace{0pt}
    \begin{subfigure}[t]{\textwidth}
      \centering
      \includegraphics[height=0.4\textheight]{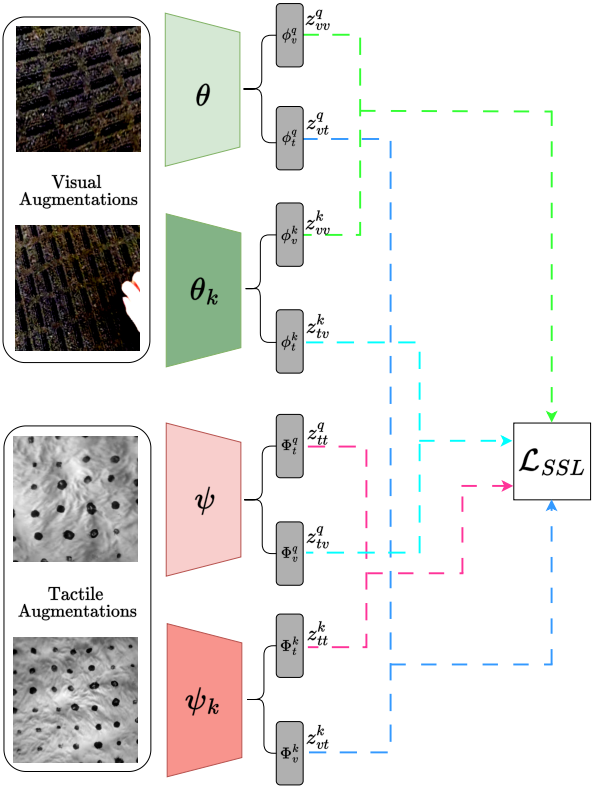}
       \tikz[remember picture] \node[coordinate] (start) {};
    \end{subfigure}
  \end{minipage}
  \begin{minipage}[t]{0.02\textwidth}
    \vspace{0pt}
    \centering
    \hspace*{\fill}\rule{0.5pt}{0.4\textheight}\hspace*{\fill}
  \end{minipage}
  \hfill
  \begin{minipage}[t]{0.50\textwidth}
    \vspace{0pt}
    \begin{subfigure}[t]{\textwidth}
      \centering
      \includegraphics[height=0.2\textheight]{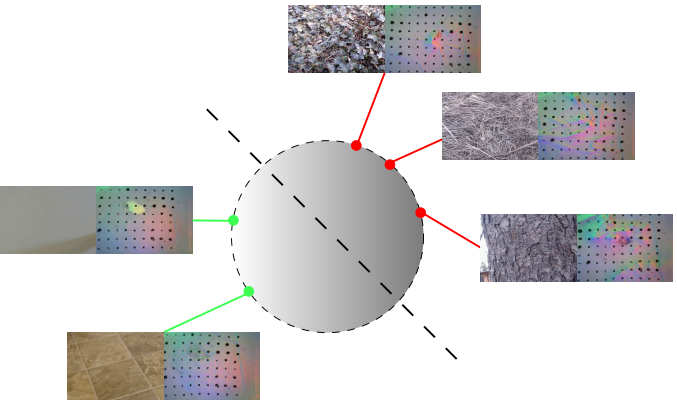}
    \end{subfigure}
    \\
    \begin{subfigure}[t]{\textwidth}
      \centering
      \includegraphics[height=0.19\textheight,width=1\textwidth]{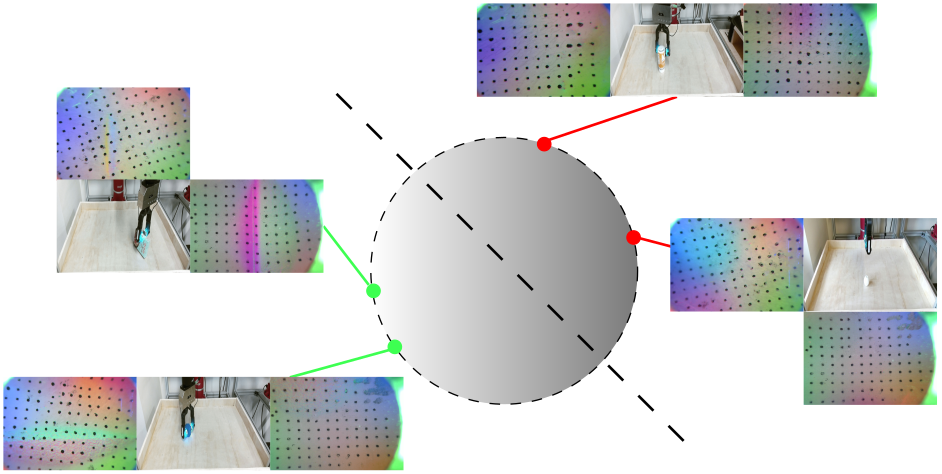}
    \end{subfigure}
  \end{minipage}
  \caption{Left: Architecture of MViTaC that consists of two training mechanisms, the inter-modal (\textcolor{green}{green} and \textcolor{pink}{pink}) and intra-modal contrastive learning (\textcolor{blue}{blue} and \textcolor{cyan}{cyan}), which collectively contribute to the self-supervised learning loss. Right: The SSL pre-trained model is subsequently evaluated on downstream tasks via linear probing. The task we consider are material property classification (top) and grasp success prediction (bottom).}
  \label{fig1:mmssl}
\end{figure*}

\subsection{Multimodal Learning}
Our approach is built upon Multimodal Contrastive Training (MCT)~\cite{yuan2021multimodal}, a method originally developed for learning relationships between visual and textual data. The overall adapted architecture of the proposed approach is illustrated in Figure~\ref{fig1:mmssl}. Within this framework, we utilize dual encoders: one for processing visual data and another for handling tactile information.
The training is comprised of two distinct contrastive training paradigms: intra-modal and inter-modal learning. Intra-modal refers to the learning that occurs exclusively within the same data modality, either between tactile-to-tactile or visual-to-visual representations. This loss is designed to maintain the similarity among augmented images while capturing their inherent structures, thereby enhancing the model's robustness to varying perspectives (Details regarding the augmentations are in the Appendix). Inter-modal refers to the learning that incorporates multiple data modalities, facilitating the understanding of relationships between different types of data, such as tactile and visual representations. 

\subsection{Architecture}
Our dataset, comprised of $N$ samples, is formally defined as $\mathcal{D} = \{(v_i, t_i)\}_{i=1}^N$, where the tuple $(v_i,t_i)$ corresponds to an image-tactile pair.  Referring to the architecture illustrated in Figure~\ref{fig1:mmssl}, the image encoder and its momentum-based counterpart are convolutional neural networks, parameterized by $\theta$ and $\theta_k$ respectively. These encoders are built upon a ResNet-18 backbone~\cite{resnet}, pre-trained on ImageNet~\cite{imagenet}. We additionally use a 2-layer MLP Projection head, which serves to map the representations into an embedding space for contrastive estimation. For the image encoder and its momentum counterpart, two distinct MLPs are utilized to generate latent representations tailored for intra-modal and inter-modal learning tasks. The momentum encoder as well as the MLP projection heads $(\phi_{v}^k,\phi_{t}^k)$ are updated with the momentum update~\cite{moco}:
\begin{align}
\theta_k   &\leftarrow m\theta_k+(1-m)\theta, \\
\phi_{v}^k &\leftarrow m\phi_{v}^{k}+(1-m)\phi_{v}^{q}, \\
\phi_{t}^k &\leftarrow m\phi_{t}^{k}+(1-m)\phi_{t}^{q}.
\end{align}
Regarding the tactile models, the encoder and its momentum counterpart are instantiated with identical configurations and are utilized to ensure consistency in the encoding process.

\subsection{Intra-modal learning}
Intra-modal learning aims at optimizing the feature representations within the same sensory modality, such as visual-to-visual or tactile-to-tactile, to enhance the discriminative power of the extracted features and the generalization capabilities of the model for downstream tasks. One common practice in contrastive learning is to randomly augment a single image to obtain different representations~\cite{simCLR}, which are then passed to the encoder to obtain a representation called ``query" and its momentum counterpart to obtain a``key"~\cite{moco}. We derive two encodings for each modality from their respective inputs: for visual images $\{o_v^q,o_v^k\}$ from $o_v$ and for tactile image $\{o_t^q,o_t^k\}$ from $o_t$. The encodings for visual data can be written as follows
\begin{align}
  z_{vv}^q &= g(f(o_v^q;\theta);\phi_v^q)\lowcomma \\  
  z_{vv}^k &= g(f(o_v^k;\theta_k);\phi_v^k)\lowcomma
\end{align}
where $\phi_v^q$ and $\phi_v^k$ are the query and key projection heads for visual data respectively and $z_{vv}^q$ and $z_{vv}^k$  are their corresponding visual embeddings generated by their respective query and key encoders (represented as $f(\cdot)$) in conjunction with their associated projection heads (represented as $g(\cdot)$).\newline
Similarly, we obtain encodings for the tactile images as
\begin{align}
  z_{tt}^q &= g(f(o_t^q;\psi);\Phi_t^q)\lowcomma \\  
  z_{tt}^k &= g(f(o_t^k;\psi_k);\Phi_t^k)\lowdot
\end{align}

We adopt InfoNCE~\cite{CPC} loss to optimize the contrastive learning objective. Given a query $z_{vv}^q$ with a positive sample $z_{vv}^{k+}$ and K negative samples denoted by $z_{vv}^{k_i}$ with $k_i=\{1...K\}$, the InfoNCE loss can be defined as
\begin{equation}
\label{eq5:infonce_vv}
\mathcal{L}_{vv} = - \log \frac{\exp(z_{vv}^q \cdot z_{vv}^{k^+} / \tau)}{\sum_{i=0}^{K} \exp(z_{vv}^q \cdot z_{vv}^{k_i} / \tau)}.
\end{equation}
Here $\tau$ is the temperature hyper-parameter~\cite{wu2018unsupervised} for $\ell_2$-normalized q and k. In contrast to MoCo~\cite{moco} and MCT~\cite{yuan2021multimodal}, which employs a memory queue to maintain negative samples, our approach aligns with previous works~\cite{simCLR,ye2019unsupervised,chen2021empirical}, utilizing the other keys present in the same batch as negative samples. In our model, the intra-modal learning consists of two parts, the visual-to-visual $\mathcal{L}_{vv}$ as Eq. (\ref{eq5:infonce_vv}) and the tactile-to-tactile contrastive learning, formalised as $\mathcal{L}_{tt}$ loss and in a similar manner as Eq.~$(\ref{eq5:infonce_vv})$.

\subsection{Inter-modal learning}
Inter-modal learning aims at aligning the feature spaces across the visual and touch sensory modalities, in order to develop a unified multimodal latent representation that captures the underlying relationships and shared attributes between these distinct data sources.

\subsubsection{Image-to-Tactile representation learning}
For a visual image $o_v$ and its corresponding tactile image $o_t$, we utilize the image encoder to produce the query feature and the momentum tactile encoder for the generation of the key feature. They are then projected to latent space by their respective projection heads
\begin{align}
  z_{vt}^q &= g(f(o_v^q;\theta);\phi_t^q)\lowcomma \\  
  z_{vt}^k &= g(f(o_t^k;\psi_k);\Phi_v^k)\lowdot
\end{align}
Contrary to the approach employed by MCT~\cite{yuan2021multimodal},  which utilizes dot products between learned representations and introduces a `margin" hyper-parameter, our method adopts the InfoNCE loss as specified in Eq.~(\ref{eq5:infonce_vv}). This deviation arises as we operate in the image space for both modalities, a distinction from the scenario in~\cite{yuan2021multimodal} that handles textual data instead. The corresponding loss $\mathcal{L}_{vt}$ is also defined as InfoNCE loss and is formulated in the same manner as in Eq.~(\ref{eq5:infonce_vv})
\begin{equation}
\label{eq8:infonce_vt}
\mathcal{L}_{vt} = - \log \frac{\exp(z_{vt}^q \cdot z_{vt}^{k^+} / \tau)}{\sum_{i=0}^{K} \exp(z_{vt}^q \cdot z_{vt}^{k_i} / \tau)}
\end{equation}

\subsubsection{Tactile-to-Image representation learning}
Similar to Image-to-Tactile learning, given a visual-tactile image pair ${o_v,o_t}$, we generate the query feature from the tactile encoder and the key feature from the momentum image encoder, which are subsequently projected to the latent space by their respective projection heads as
\begin{align}
  z_{tv}^q &= g(f(o_t^q;\psi);\Phi_v^q) \\  
  z_{tv}^k &= g(f(o_v^k;\theta_k);\phi_t^k)
\end{align}
The loss $\mathcal{L}_{tv}$ is also formalised as InfoNCE loss and is formulated in the same manner as in Eq.~(\ref{eq8:infonce_vt}).

\subsection{Combined Loss}
The overall multimodal contrastive loss used in MViTac is formulated as a combination of both, inter-modal and intra-modal contrastive loss, weighted by a term $\lambda_{inter}$ that is responsible for the trade-off between between the optimization objectives for inter-modal and intra-modal representations. The overall loss $\mathcal{L}_{mm}$ is defined as
\begin{align}
\mathcal{L}_{mm} = \mathcal{L}_{vv} + \mathcal{L}_{tt} + \lambda_{inter} (\mathcal{L}_{vt} + \mathcal{L}_{tv})
\end{align}

\section{EXPERIMENTS}
In this section, we discuss the experiments aimed at validating the performance of our proposed MViTac learning framework. We elaborate on the two datasets used and evaluate the learned representations on four downstream tasks. Our model is benchmarked against state-of-the-art self-supervised methods in visuotactile learning and supervised learning method.

\begin{table*}[ht]  
\centering
\renewcommand{\arraystretch}{1.2}  
\caption{Comparison on different material property identification downstream tasks}
\begin{tabular}{l@{\hspace{1.5em}} c@{\hspace{1.5em}} c@{\hspace{1.5em}} c@{\hspace{1.5em}} c@{\hspace{1.5em}} c}
\hline
\textbf{Dataset} & \textbf{Method} & \textbf{Modality} & \makecell{\textbf{Category} \\ \textbf{Accuracy\%}} & \makecell{\textbf{Hard/Soft} \\ \textbf{Accuracy (\%)}} & \makecell{\textbf{Rough/Smooth} \\ \textbf{Accuracy (\%)}} \\ \hline
Chance                      & - & Tactile                           & 18.6 & 66.1         & 56.3 \\ 
ResNet18~\cite{resnet}          & Supervised Learning & Tactile    & 57.4 & \textbf{\textcolor{blue}{89.1}} & 79.3 \\ 
                                &    & Tactile + Visual  & 48.0 & 85.9 & 80.0 \\ 
TAG~\cite{yang2022touch}        & Contrastive Multiview Coding & Tactile& 54.7 & 77.3 & 79.4 \\ 
                                &              & Tactile + Visual  & 68.6 & 87.1 & 82.4 \\ 
SSVTP~\cite{kerr2023selfsupervised} & InfoNCE & Tactile           & 46.1 & 79.7 & 75.8 \\
                                    &  & Tactile + Visual  & 70.7 & 88.6 & 83.6 \\ \hdashline
MViTac (Ours)   & Multimodal Contrastive Training & Tactile & \textbf{\textcolor{blue}{57.6}} & 86.2 & \textbf{\textcolor{blue}{82.1}} \\ 
                &  & Tactile + Visual & \textbf{\textcolor{red}{74.9}} & \textbf{\textcolor{red}{91.8}} & \textbf{\textcolor{red}{84.1}} \\ \hline
\end{tabular}
\label{table1:classification}
\caption*{We report the evaluation of Top-1\% accuracy across various downstream tasks. We evaluate on both, tactile-only and visual-tactile data. The \textcolor{blue}{blue} are the best results in tactile-only modality and \textcolor{red}{red} shows the best result in tactile+visual modality.}
\end{table*}

\subsection{Experimental Setup}
\subsubsection{Material property identification} We evaluate on the Touch-and-Go (TAG)~\cite{yang2022touch} dataset to solve the task of material property identification. This dataset encompasses a diverse range of tactile features that are instrumental in bifurcating various material properties. We consider three downstream tasks: 1) categorization of materials, 2) distinction between hard and soft surfaces, and 3) distinction between smooth and textured surfaces. We adhere to the dataset splits prescribed by the authors of ~\cite{yang2022touch} to maintain experimental consistency. This partitioning ensures that our evaluations are directly comparable to prior work and their baselines. For the classification task, the dataset comprises a collection of 20 distinct objects, whereas Hard/Soft and Rough/Smooth are binary classification task.

\subsubsection{Robot Grasping Prediction} The Calandra dataset \cite{Calandra2018MoreTA} provides the data from a pair of tactile sensors attached to a jaw gripper (left and right) alongside the RGB images. A triplet of samples was captured 'before', 'during', and 'after' grasping a plethora of objects. The objective is to determine the success or the failure of the grasp attempt. Unlike in TAG \cite{yang2022touch}, the Calandra dataset does not provide a predefined train/test split, leading us to create our own randomized split. We train our model on a subset of 40 unique objects from the total 106 of the original dataset, keeping only the demonstrations with the most grasping attempts. In contrast to TAG, which only utilizes tactile-tactile encoding in their experiments, we use visual-tactile representations to predict the grasp success. The only difference with the setup in the previous section is that we stack the tactile images from the two sensors across the channel dimension before passing them through the respective encoder to obtain their common representation. Finally, we evaluate the learned representations by using the tactile pair and the RGB image from the 'during' samples to train a linear classifier.

\subsection{Evaluation and Results}
We evaluate our MViTac model against established models on tactile-only and visual-tactile data. For reproducibility and further research, we provide our test/train split of the Calandra dataset, along with the models and code, on our project website\footnote{
\begin{sloppy}
https://sites.google.com/view/mvitac/home
\end{sloppy}
}.

\subsubsection{Material property identification}
We first evaluate MViTac on only the tactile data for all the three material property identification tasks.
For a comparative assessment, we employed multiple methodologies within the same problem domain.
These encompass a supervised learning framework that relies on ResNet-18 architecture~\cite{resnet}, where the model is trained on labeled data. Furthermore, we incorporate comparative metrics from TAG~\cite{yang2022touch}, a model that utilizes the Contrastive Multiview Coding (CMC) approach~\cite{tian2020contrastive} to learn cross-modal representations. The baseline results for TAG are sourced from the original paper as the experimental conditions, including data splits, and were maintained in strict accordance with the original study. In addition, we also consider SSVTP~\cite{kerr2023selfsupervised}, a recent approach that employs InfoNCE~\cite{CPC} loss in its pre-training phase for self-supervised learning.\newline
In accordance with common practice, we evaluate the quality of the learned representations through linear probing. Following the self-supervised pre-training phase, we detach the projection heads and freeze the encoder. On these static representations, we then train a linear classifier in a supervised fashion.\newline
As observed in Table~\ref{table1:classification}, the tactile-only model surpasses the performance of both the cross-modal representation learning methods, TAG and SSTVP. Notably, our method exceeds the performance of not just self-supervised approaches like TAG and SSTVP, but also the supervised learning model, which commonly surpasses self-supervised methods in general efficacy. Nonetheless, the supervised learning approach exhibits a marginal advantage in tasks related to the classification of hard and soft surfaces. Remarkably, the integration of visual and tactile data results in substantial performance gains. Across all methodologies, incorporating visual information appears to uniformly elevate prediction accuracy—an outcome that, while expected, underscores its significance.\newline
Within this framework, our method consistently outperforms all alternative approaches, signifying the effectiveness of our multimodal strategy. Although both TAG and SSVTP leverage contrastive learning frameworks and operate on closely related loss functions, their performance lags behind our model in terms of prediction accuracy. Even though TAG and SSVTP are contrastive learning models, and work on almost similar loss functions, they achieve less prediction accuracy than our model. Our model not only distinguishes similarities and differences between visual and tactile data but also places substantial emphasis on learning within the same sensory modality.

\begin{table}[h]  
\centering
\renewcommand{\arraystretch}{1.2}  
\caption{Comparison of predicting the success of grasping}
\begin{tabular}{l@{\hspace{1em}} c@{\hspace{0em}} c}
\hline
\textbf{Dataset} & \textbf{Method} & \makecell{\textbf{Grasping Pred.} \\ \textbf{Accuracy\%}} \\ \hline
TAG~\cite{yang2022touch}                        & Contrastive Multiview Coding    & 56.3 \\
MViTac (Ours)                                   & Multimodal Contrastive Training    & 60.3 \\ \hdashline
Calandra et. al.~\cite{Calandra2018MoreTA}      & Supervised Learning (Baseline)     & \textbf{73.1} \\ \hline
\end{tabular}
\caption*{Top 1\% Accuracy. The best result is in bold.}
\end{table}

\subsubsection{Robot Grasping Prediction}
We compare the robustness of the learned representations using again the common linear probing method and train a linear classifier to predict robot grasping outcomes. We compare our approach against the CMC method~\cite{yang2022touch} and the supervised method proposed in \cite{Calandra2018MoreTA} for solving this classification task. We report the comparison results in Table \ref{table1:classification}. The superiority of the learned embeddings using MViTac is apparent, where we outperform CMC by almost $4\%$ on predicting the grasping of unseen objects. We must note here, that the supervised method of \cite{Calandra2018MoreTA} greatly outperforms both self-supervised modalities. This discrepancy in performance is expected due to the small size of the training dataset which is around 18000 samples and its imbalanced distribution of samples across objects. CL techniques necessitate bigger and more diverse datasets to perform comparably with supervised methods.

\section{DISCUSSION AND LIMITATIONS}
We have demonstrated that our model exhibits better generalizability across real world datasets. For the material property classification, we report considerable performance improvements when visual and tactile data are combined. The results suggest that while tactile information is adept at capturing fine-grained material properties, it provides insufficient information for the accurate classification of complex surfaces, such as those encountered in uncontrolled environments. While we acknowledge that self-supervised approaches may not surpass supervised methods in scenarios with limited data, as evidenced in our grasp prediction experiment, efforts are underway to narrow this performance gap. In the grasp prediction task, we acknowledge that self-supervised methods do not yet outperform supervised methods as data is limited; however, efforts are underway to minimize this performance disparity. Our evaluations leverage datasets that, albeit comprehensive, might not fully encapsulate the complexity and variability seen in real-world scenarios. Extending the evaluation to real robotic systems and understanding how well the model performs in real-time tasks is an essential next step. Expanding the scope of evaluation to include a broader spectrum of tasks, including more sophisticated manipulation tasks, could provide a more comprehensive understanding of the model's capabilities and limitations. Nonetheless, it is important to note that our experiments are based on data collected from real-world tasks, thus affirming the relevance and value of our findings.

\section{CONCLUSIONS}
In this paper, we presented MViTac, a novel approach for incorporating both visual and tactile sensory for various tasks. More specifically, our methodology learns inter-modal and intra-modal representations via self-supervised learning which leads to more efficient representations. As a consequence, MViTac consistently surpasses existing self-supervised methodologies across all the benchmarks and outperforms supervised learning approaches on material property recognition. Furthermore, it outperforms self-supervised state-of-the-art method on grasping success prediction with linear probing. Nonetheless, it necessitates further validations on real robotic platforms to ensure its efficacy in real-world applications. Our future works will aim at refining the learning architecture and venturing into real robot experiments to foster advancements in multimodal robotic tasks.

\section*{APPENDIX}
\subsection{Implementation Details}
For image and tactile modality, we use ResNet-18~\cite{resnet} as the backbone. In order to obtain representations, we apply average pooling on the last layer of the backbone. For all the encoders i.e. for inter-modal and intra-modal contrastive learning, we use a 2-layer MLP  as projection heads that converts the 512-dimensional output from the backbone into a 128-dimensional final representation. We use ReLU~\cite{nair2010rectified} as the activation function for the first layer and no activation function for the final layer. All the representation vectors are normalised before calculating the contrastive loss. We set the temperature $\tau$ as $\{0.07,0.2,0.5,1\}$. The best results were obtained when $\tau$ was 0.07.\newline
All the networks are trained jointly using ADAM~\cite{kingma2014adam} optimizer with parameters $\beta_1 = 0.9, \beta_2=0.999,\epsilon_{ADAM}=10^{-7}$. During the pre-training phase, a learning rate of 0.03 is employed, while for the subsequent downstream tasks, a reduced learning rate of $10^{-4}$ is used. To mitigate overfitting, a dropout layer is integrated into the classifier, featuring a dropout probability of 0.2, serving as regularization. We use the batch size of 256 training on a single 4090 GPU for 240 epochs for pre-training and 60 epochs for the downstream tasks. For the downstream tasks, the projection heads~\cite{simCLR}, are bypassed; instead, the encoder's output is directly fed into the input layer of the classifier.

\subsection{Augmentations}
We follow the standard practice of resizing the images into 256x256 pixels and subsequently normalized using predefined mean and standard deviation metrics. Then the images are subjected to a randomly resized crop with dimensions of 224x224 pixels, falling within a scale range of 0.2 to 1.0. Additional stochastic operations include the application of horizontal flipping with a 50\% probability and the conversion to grayscale with a 20\% probability. It should be noted that the grayscale transformation is excluded from the grasping task, as it offers little utility in such a controlled experimental setting.

\subsection{Baseline Methods}
In the case of supervised learning, we employ a pair of ResNet-18 encoders~\cite{resnet} that independently generate tactile and visual representations, which are subsequently concatenated and directly subjected to the classification process. This strategy diverges from our other baselines, where we employ a linear classifier post-encoding for task-specific adaptations. 
In our study, we adopt a feature dimension of 128 and a batch size of 128, diverging from SSVTP's settings of 8 and 512 respectively, due to dataset size and computational constraints~\cite{kerr2023selfsupervised}. In our study, the TAG model~\cite{yang2022touch} was executed as per the original specifications, leveraging the code made publicly available by the authors. Encoders in all the baselines are built upon a ResNet-18 backbone~\cite{resnet}, pre-trained on ImageNet~\cite{imagenet}, with a modification in the final layer. For the grasping experiment~\cite{Calandra2018MoreTA}, we employed the implementation available in Tacto~\cite{Wang2022TACTO}.

\section*{ACKNOWLEDGMENT}
This project has received funding from the Deutsche Forschungsgemeinschaft (DFG, German Research Foundation) - No \#430054590 (TRAIN).
\clearpage

\printbibliography

\end{document}